\documentclass[letterpaper]{article} 
\usepackage{aaai24}  
\usepackage{times}  
\usepackage{helvet}  
\usepackage{courier}  
\usepackage[hyphens]{url}  
\usepackage{graphicx} 
\usepackage{natbib}  
\usepackage{caption} 
\usepackage{algorithm}
\usepackage{algorithmic}

\usepackage{newfloat}
\usepackage{listings}
\DeclareCaptionStyle{ruled}{labelfont=normalfont,labelsep=colon,strut=off} 
\floatstyle{ruled}
\newfloat{listing}{tb}{lst}{}
\floatname{listing}{Listing}
\title{WC-SBERT: Zero-Shot Text Classification via SBERT with Self-Training for Wikipedia Categories}
\author{
    Te-Yu Chi\equalcontrib\textsuperscript{\rm 1},
    Yu-Meng Tang\equalcontrib\textsuperscript{\rm 1,\rm 2},
    Chia-Wen Lu\equalcontrib\textsuperscript{\rm 1},
    Qiu-Xia Zhang\equalcontrib\textsuperscript{\rm 1},
    Jyh-Shing Roger Jang\textsuperscript{\rm 1}
}
\affiliations{
    \textsuperscript{\rm 1}National Taiwan University\\
    \textsuperscript{\rm 2}Tongji University\\
    d09922009@ntu.edu.tw, 
    tonmoregulus@gmail.com,
    b07610058@ntu.edu.tw,
    r10922164@ntu.edu.tw,
    jang@mirlab.org

}

\begin{document}

\maketitle

\begin{abstract}
In the research domain of Natural Language Processing (NLP), the problems of intent and topic classification have consistently been fundamental and significant. These issues demand that machines comprehend the contextual meaning of a sentence, thereby addressing various language-related problems such as Question-Answering (QA) and Multi-dialogue. In many classification tasks, it is customary to rely on training datasets to enhance the model's accuracy. However, in practical business applications, it is sometimes challenging to gather a sufficient amount of data for training purposes.

Our research focuses on solving the zero-shot text classification problem in NLP, with a particular emphasis on innovative self-training strategies. To achieve this objective, we propose a novel self-training strategy that uses labels rather than text for training, significantly reducing the model's training time. Specifically, we use categories from Wikipedia as our training set and leverage the SBERT pre-trained model to establish positive correlations between pairs of categories within the same text, facilitating associative training.

For new test datasets, we have improved the original self-training approach, eliminating the need for prior training and testing data from each target dataset. Instead, we adopt Wikipedia as a unified training dataset to better approximate the zero-shot scenario. This modification allows for rapid fine-tuning and inference across different datasets, greatly reducing the time required for self-training.

Our experimental results demonstrate that this method can adapt the model to the target dataset within minutes. Compared to other BERT-based transformer models, our approach significantly reduces the amount of training data by training only on labels, not the actual text, and greatly improves training efficiency by utilizing a unified training set. Additionally, our method achieves state-of-the-art results on both the Yahoo Topic and AG News datasets.
\end{abstract}

\section{Introduction}

With the rapid advancement of Natural Language Processing (NLP) technologies, zero-shot text classification has become a hot research topic in this field. Zero-shot text classification is a unique text classification task, distinguished by its ability to classify text into predefined categories without any prior labeled data. This problem holds significant research value in NLP as it not only copes with known categories but also deals with newly appearing categories and texts from unknown domains, showcasing a remarkable generalization capability.

Despite the recent success achieved by \citet{selftrain} in using a self-training approach, wherein prediction results were used as labels on an unlabeled dataset, and these pseudo-labels were subsequently used to train the model through multiple iterations, this method still leaves room for further improvement and optimization.

This paper presents a novel self-training strategy, which focuses on training on labels rather than texts. This method not only reduces model inference time but also improves training efficiency. Our proposed strategy utilizes the SBERT \citep{reimers-2019-sentence-bert} model for sentence-level vector representation to train on labels by establishing their positive associations, rather than training on texts. During the self-training phase, we can quickly perform a large number of inferences, greatly reducing the inference time.

Our method was validated on the Yahoo Topic and AG News datasets and compared with existing self-training methods. Our experimental results show that by self-training on labels, our method can adjust a model to a target dataset in a matter of minutes, and compared to other BERT-based transformer models, our method can drastically reduce the volume of training data, thereby enhancing training efficiency.

We also compared our methods with the popular GPT and GPT fine-tune methods in the experiments. Directly using GPT for text classification is cost-effective but only provides average accuracy. On the other hand, fine-tuning on the superior Curie model provides preliminary accuracy close to the state of the art, but it's expensive and unsuitable for inference on large-scale datasets. In contrast, our model is more effective and requires fewer computational resources.

Therefore, our model has several contributions:

\begin{enumerate}
  \item It achieves state-of-the-art results on the Yahoo Topic and AG News datasets.
  \item Compared to self-training methods by \citeauthor{selftrain}, it does not require preliminary self-training on the target dataset's training set. Instead, it uses Wikipedia data for self-training. This allows universal application across datasets, greatly reducing self-training inference time by encoding Wikipedia data in advance and storing the embeddings in H5 format.
  \item Compared to other BERT-based transformer models that primarily train on texts, our study trains only on categories (labels), significantly reducing the amount of training data, thereby enhancing training efficiency and reducing computational resource consumption.
  
\end{enumerate}

The remaining part of this paper will further detail our self-training strategy, including the selection of the SBERT model, why we view Wikipedia page categories as labels, and how we fine-tune the model using the self-training strategy. Moreover, we will explain how we applied this method to our experiments, designed them to effectively validate our approach, and discuss our research findings and potential applications and limitations of our method. In summary, we hope this study can provide a novel, effective solution to the zero-shot intent classification problem. We also aim to stimulate more discussion and research on this topic.

\section{Related work}
Zero-shot text classification is a text classification task that has the characteristic of classifying text into predefined classes without prior labeled data. We focus on open-domain zero-shot text classification, which has better generalization capability compared to specific domains. It can not only classify text into known classes but also handle new classes and texts from unknown domains. \citet{chang2008} conducted the initial research on this task, referred to as "dataless classification" at that time. It is a method that relies solely on general knowledge for classification and does not require any domain-specific data. Since then, this task has gained significant attention.

With the advancement of deep neural networks, text classification has experienced a significant shift towards the use of pre-trained language models (PLMs) (\citeauthor{plm1}, \citeyear{plm1}; \citeauthor{plm2}, \citeyear{plm2}; \citeauthor{plm3}, \citeyear{plm3}). The first-generation PLMs, such as Word2Vec \citep{word2vec} and GloVe \citep{glove}, relied on word embeddings and typically classified the text based on word similarity. While these models were effective in capturing the semantic meaning of words and sentences, they lacked the ability to understand complex linguistic concepts and contextual information. In contrast, the second-generation PLMs, such as BERT \citep{bert2019}, GPT-2 \citep{glove}, and RoBERTa \citep{roberta}, are based on contextual embeddings. These models become mainstream techniques in text classification since they can capture the semantic meaning of words in different contexts and can be fine-tuned for this NLP task.

Recently, \citet{tewiki} as well as \citet{chu2020} both utilized Wikipedia data as training data to construct BERT-based classifiers. This choice is due to the vast amount of article data available in Wikipedia, which covers a wide range of general knowledge. Therefore, it is considered highly suitable for open-domain zero-shot text classification tasks. The datasets constructed by \citeauthor{chu2020} and \citeauthor{tewiki} contain 5.75 million documents and 1.19 million categories, and 3.3 million documents and 674 top-level categories, respectively. Compared to their works, we select Wikipedia dataset from Huggingface as data source. The dataset stands out for its substantial size, comprehensive coverage, and meticulous categorization. With over 6 million documents and more than 1.5 million categories, it has shown better performance on popular text classification datasets in our experiments.

Self-training is one of the commonly used techniques in the fields of semi-supervised \citep{semi} and unsupervised learning \citep{unsuper}. It is characterized by utilizing model predictions to expand the training dataset and iteratively improve the model through self-training. In recent studies (\citeauthor{zeroshot1}, \citeyear{zeroshot1}; \citeauthor{zeroshot2}, \citeyear{zeroshot2}; \citeauthor{zeroshot3}, \citeyear{zeroshot3}; \citeauthor{selftrain}, \citeyear{selftrain}) about self-training in zero-shot text classification, \citet{selftrain} overall achieved the best performance. Self-training involves pseudo-labeling the training and test sets of the target dataset. Subsequently, the model undergoes multiple iterations of training using these pseudo-labels to enhance its understanding of the dataset. While this approach has indeed demonstrated state-of-the-art results on multiple datasets, it requires prior examination of the data in each individual target dataset. In certain commercial contexts, this may not accurately represent the absence of zero-shot data, necessitating additional computational resources to train the model on different datasets. This study aims to improve the self-training approach in terms of dataset selection and performance.

\section{Methodology and approach}
Previous research has extensively investigated various methods and models for zero-shot intent classification problems. These approaches encompass BERT-based \citep{tewiki} methods that involve training on text and categories, as well as the utilization of generative models such as GPT \citep{gpt-zero} and T5 \citep{t5-zero} to guide the generation of desired answers. Among these approaches, the commonly adopted strategy involves training BERT on text and categories and utilizing entailment to ascertain the relationship between hypotheses and premises, classifying them as entailment, neutral, or contradiction. However, utilizing this approach requires training on the given text, which can be time-consuming.

Recent studies have employed a self-training approach \citep{selftrain}, wherein the original model undergoes iterative inference on the training set of the target dataset. Data instances surpassing a predetermined confidence threshold are then employed as training data for subsequent fine-tuning. Although this approach attains state-of-the-art performance in intent classification, it presents two challenges:

\begin{itemize}
    \item Acquiring pre-training data from diverse target datasets for the purpose of fine-tuning.
    \item Addressing the first challenge, performing inference on different datasets necessitates relatively more time for processing encoding embeddings. Notably, embedding processing constitutes one of the most time-consuming steps in BERT processing.
\end{itemize}

To overcome these aforementioned challenges, this study introduces a methodology that achieves state-of-the-art performance in zero-shot classification while resolving the aforementioned issues. The subsequent section outlines the methodology employed in this research.

\subsection{SBERT}
SBERT (\textit{Sentence-BERT}) \citep{reimers-2019-sentence-bert} is a method based on BERT (\textit{Bidirectional Encoder Representations from Transformers}) that calculates sentence-level vector representations. It utilizes a siamese architecture consisting of two BERT networks with shared weights, where each network processes an input sentence. These two BERT networks learn representations of the two sentences through parameter sharing, enabling comparison of sentence similarity using methods such as cosine similarity. Furthermore, in the fine-tuning process of SBERT for different downstream tasks, the choice of the loss function is crucial. In this study, we employ the multiple ranking loss (MNR Loss), which is particularly suitable when the training dataset contains only positive pairs. The MNR Loss brings positive data closer together in the vector space and pushes negative data further apart, thus forming clusters of similar sentences. Leveraging the architecture and training/inference optimizations of SBERT compared to traditional BERT, this study utilizes the pre-trained model \textit{all-mpnet-base-v2} from SBERT as the primary base pre-trained model.

We introduces a novel training approach for BERT-based models, shifting from the conventional training method to an association training based on model labels (categories). The purpose of the basic model fine-tuning stage is to fine-tune the SBERT-based model using the Wikipedia dataset to construct training data and obtain a generalized base model. Based on the provided URLs in the dataset, we identify the corresponding category for each data entry's text. Each data entry in the dataset is defined by three components: id \(i\), text \(t_i\), and the set \(C_i\) of categories to which the text belongs. Assume that $C_i = \left\{c_1, c_2,...,c_{n_i}  \right\}$, $n_i$ represents the size of the category set for the \(i^{th}\) data entry. We use each category in the set $C_i$ for pairwise combinations. For each data entry, we can obtain \(C^{n_i}_{2}\) pairs of classes as training data. The set \(P_i\) comprises pairs of classes for each data entry. Its definition is shown below. We select all \(P_i\) where $n_i \geq 2$ to construct the initial training set and perform fine-tuning on the pre-trained model. 

\begin{quote}
    $P_i = \left\{(c_j, c_k) | c_j, c_k \in C_i, c_j \neq c_k \right\}$
\end{quote}

Regarding the dataset, we utilize the Wikipedia dataset from Huggingface and assume that the content of each wiki, along with its associated categories, exhibits positive correlations based on their respective wiki page IDs. As a result, we modify the training approach of most BERT-based models for the given text and adopt associated training using model labels (categories). Since the length of labels is significantly shorter compared to the content of the text, this reduces training costs. We use the pre-trained SBERT model as the base model and treat the categories of each wiki page in \textit{wiki-cate} as labels. We combine these labels in pairs (non-repetitive combinations using Python's \textit{itertools.combinations}) and convert them into the SBERT training data format (\textit{InputExample}) to create the training dataset \textit{train-samples}. Pseudo code is shown below:

\begin{listing}[h]%
\caption{Pseudo code of generate SBERT training samples}
\label{lst:listing}%
\begin{lstlisting}[language=Python]
import itertools
from sentence_transformers import InputExample

train_samples = []

for data in wiki_cate:
    categories = data["categories"]
    for pair in list(itertools.combinations(categories, 2)):
        train_samples.append(InputExample(texts=[pair[0], pair[1]]))
\end{lstlisting}
\end{listing}

Through SBERT, we train on the aforementioned train-samples to obtain a general purpose foundational model called \textit{wiki-cate} SBERT model: \textit{WC-SBERT}.

\subsection{Wiki text embeddings}


After the aforementioned data processing, the size of the dataset \(D\) is 6,458,670. We utilize the general base model WC-SBERT to encode the first 200 words of the text in each data point from \(D\). The resulting embeddings are stored in an h5 file for subsequent fine-tuning.

\subsection{Self-training}


In the self-training fine-tune stage, we define the target label set as \(L\). Initially, we use WC-SBERT to calculate the similarity between each text and all elements \(l\) in the label set \(L\). Then, for each text \(t_i\), we select the label \(l_j\) from \(L\) that has the highest similarity score with \(t_i\) to obtain the similarity score $similarity (t_i, l_j)$ between \(t_i\) and \(l_j\), and compare the similarity between them. Next, we compare the similarity with the set threshold. If the similarity is greater than the threshold, we pair all elements ($c_1, c_2,..., c_{n_i}$) of the class set \(C_i\) of \(t_i\) with the target label \(l_j\) to get $(c_1, l_j), (c_2, l_j),..., (c_{n_i}, l_j)$, all text categories-target label pairs are used as training data for the fine-tuning stage. Since we have already encoded the text and stored it in an h5 file in the previous step, during the similarity calculation stage, we only need to calculate the embeddings of the target labels, and the text embeddings can be directly retrieved from the h5 file. With the advantage of SBERT's fast similarity inference, we can quickly obtain $similarity (t_i, l_j)$ for each million data points.

\begin{figure}[h]
\centering
\includegraphics[scale=0.18]{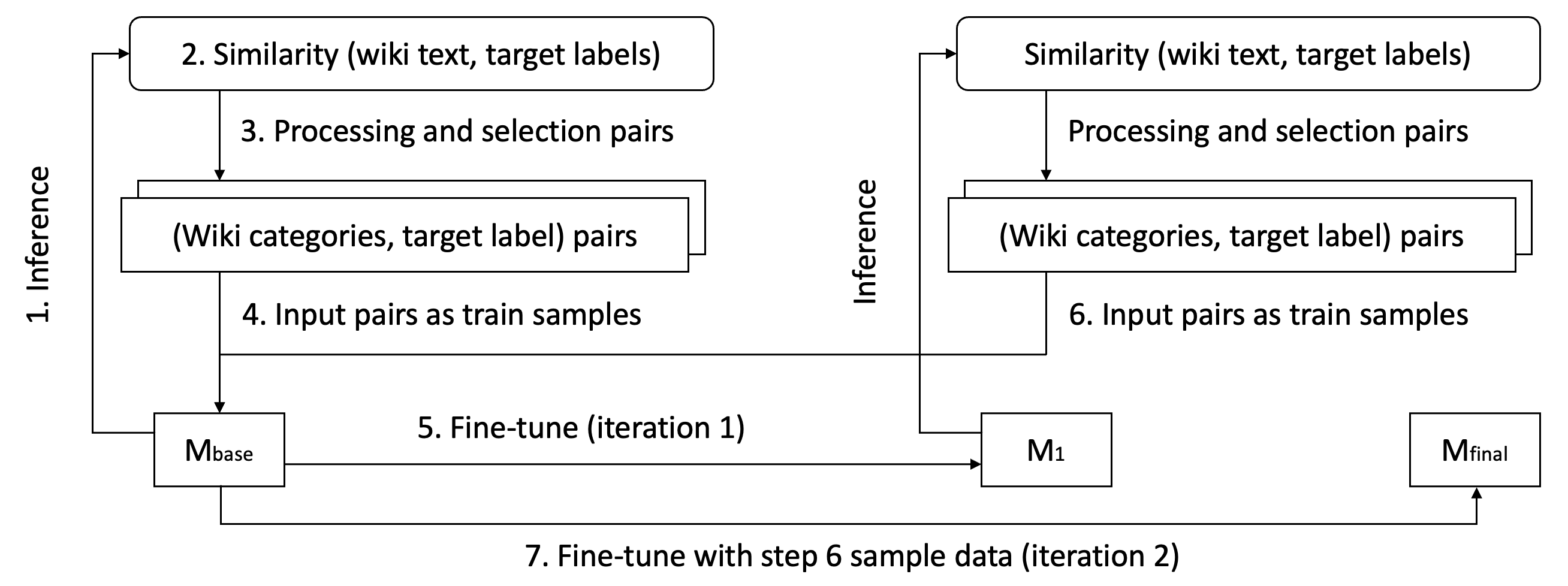}
\caption{\textbf{Fine-tune process in self-training.}}
\label{self-train_fine-tine}
\end{figure}

As shown in the Figure \ref{self-train_fine-tine}, we use the method described above to obtain the training data and fine-tune the original model \(M_{base}\), resulting in model \(M_1\). We then continue to use the same method, using \(M_1\) to calculate similarity scores and filter out the data for further fine-tuning of \(M_{base}\), obtaining model \(M_2\). We conduct this process iteratively, until we obtain the final model \(M_{final}\).

\section{Dataset}
\subsection{Training dataset}
This study utilized the train set of Wikipedia dataset from Huggingface as the training set for the pre-trained model. The dataset consists of a total of 6,417,006 records, with fields including wiki page ID, URL, title, and text. Since this dataset does not include the categories for each page, a separate web scraping program was designed in this study to retrieve the categories for each page. A total of 1,563,193 categories were collected, and the combination of these categories with the original dataset was named the \textit{wiki-cate} dataset. Here are some examples of the data:

\begin{listing}[h]%
\caption{Wikipedia sample data format}
\label{lst-wiki-data-format}%
\begin{lstlisting}
"id": 12,
"url": "https://en.wikipedia.org/wiki/Anarchism",
"title": "Anarchism",
"text": "Anarchism is a political philosophy and movement...",
"categories": ["Anarchism", "Anti-capitalism", "Anti-fascism", "..."]
\end{lstlisting}
\end{listing}

\subsection{Evaluation dataset (target dataset)}
\begin{itemize}
    \item \textbf{AG News (AG's News Corpus)}: The AG News dataset \citep{yahoo-agnews} consists of news titles and descriptions, categorized into four classes: \textit{World}, \textit{Sports}, \textit{Business}, and \textit{Sci/Tech}. This study utilizes the AG News dataset from Huggingface, with a total of 7,600 data points in the test set. \textit{Sci/Tech} is further divided into two classes (\textit{Science} and \textit{Technology}) for classification purposes.
    \item \textbf{Yahoo! Answers}: The Yahoo! Answers dataset \citep{yahoo-agnews} mainly comprises questions and answers from the Yahoo! platform. In this study, Yahoo! Answers dataset from Huggingface is employed, containing 60,000 data points in the test set. There are ten main categories for the topics: \textit{Society \& Culture}, \textit{Science \& Mathematics}, \textit{Health}, \textit{Education \& Reference}, \textit{Computers \& Internet}, \textit{Sports}, \textit{Business \& Finance}, \textit{Entertainment \& Music}, \textit{Family \& Relationships}, and \textit{Politics \& Government}. Each label is individually treated as a single label for classification (splitted by the '\&' character), and the output is mapped back to the original label. 
    \item \textbf{DBpedia}: The DBpedia dataset \citep{dbpedia} is derived from structured information extracted from Wikipedia. In this study, the DBpedia dataset \textit{dbpedia\_14} from Huggingface is used as the experimental subject, consisting of 70,000 data points in the test set. It includes 14 distinct categories, namely \textit{Company}, \textit{EducationInstitution}, \textit{Artist}, \textit{Athlete}, \textit{OfficeHolder}, \textit{MeanOfTransportation}, \textit{Building}, \textit{NaturalPlace}, \textit{Village}, \textit{Animal}, \textit{Plant}, \textit{Album}, \textit{Film}, and \textit{WrittenWork}. We will split the partial classification into words by separating them with a space as input: \textit{EducationInstitution} to \textit{Education institution}, \textit{OfficeHolder} to \textit{Office holder}, \textit{MeanOfTransportation} to \textit{Mean of transportation}, \textit{NaturalPlace} to \textit{Nature place} and \textit{WrittenWork} to \textit{Written work}.
    \label{dbpedia}
\end{itemize}

In this study, the labels of the test dataset are pre-combined with prompts to perform actual classification. Table \ref{table-ds-prompt} shows the prompts for each target dataset.

\begin{table}[h]
\centering
\begin{tabular}{l|l}
    \hline
    \textbf{Dataset} & \textbf{Prompt} \\
    \hline
    AG News & This topic is talk about \{label\}.\\
    Yahoo! Answers & This topic is talk about \{label\}.\\
    DBpedia & This sentence is belong to \{label\}.\\
    \hline
\end{tabular}
\caption{\textbf{Prompts used in the target datasets.}}
\label{table-ds-prompt}
\end{table}

\section{Experiments}
This study prioritizes conducting relevant experiments using the AG News dataset as the target dataset. The findings and techniques are then applied to other target datasets, such as Yahoo topic and DBpedia, for evaluation purposes.

\subsection{Implementation Details}
The parameters used for SBERT experiments are as follows:

\begin{itemize}
    \item Pre-trained model: all-mpnet-base-v2
    \item Train batch size: 128
    \item Sequence length (token length): 128
    \item Epoch: 1
    \item Loss function: Multiple Negative Ranking (MNR) Loss
\end{itemize}

\subsection{Prompt evaluation}
\begin{figure}[h]
\centering
\includegraphics[scale=0.35]{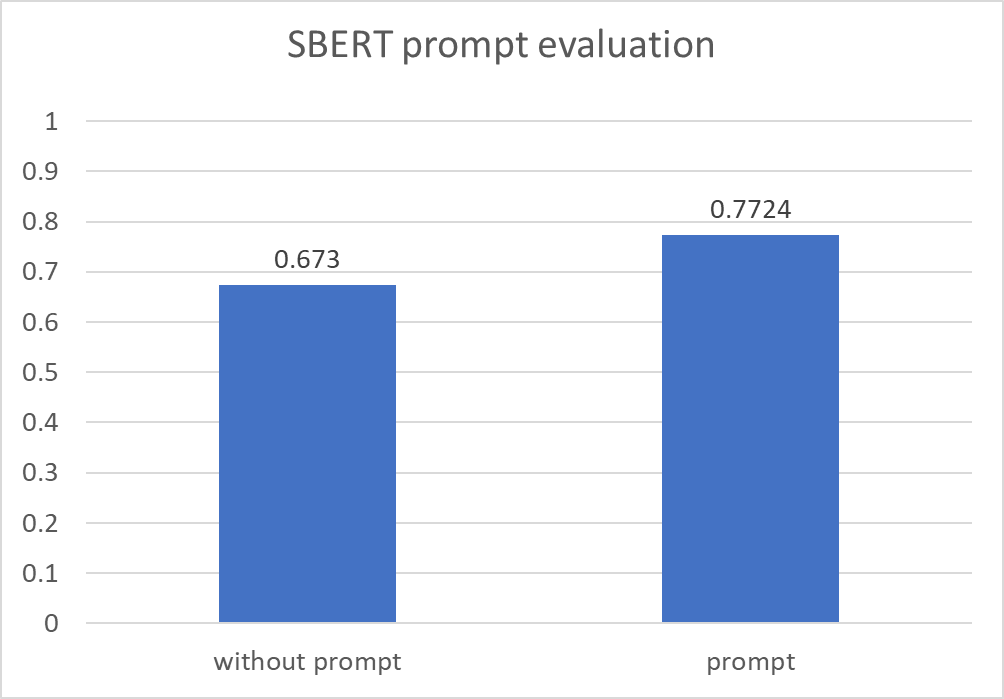}
\caption{\textbf{SBERT prompt evaluation.} The numbers denote the test accuracy of AG News. The prompt can be found in Table \ref{table-ds-prompt}.}
\label{fig-prompt-evaluation}
\end{figure}

In recent research, Prompt tuning \citep{schick2020exploiting} has shown to significantly improve the accuracy of zero-shot or few-shot classification. We aim to investigate whether Prompt tuning can also enhance the similarity matching of the SBERT-based sentence similarity model, WC-SBERT, which is trained solely on labels. In our experiments, we compare the performance of WC-SBERT using a simple hard prompt \citep{liu2022p} ("This topic is talk about [\textit{Label}]") with that of using only the label for comparison. Figure \ref{fig-prompt-evaluation} shows the evaluation results of WC-SBERT on the 7,600 test set of AG News dataset.

\subsection{Query label mapping evaluation}
WC-SBERT is pre-trained using wiki categories as labels. We are particularly interested in whether direct comparison of the target test set's queries and labels should be conducted, or if an intermediate step is required. This intermediate step involves mapping the queries to the closest category within the wiki categories and then performing the category-label matching. Intuitively, the presence of seen categories during WC-SBERT's training process should potentially improve accuracy.

\begin{figure}[h]
\centering
\includegraphics[scale=0.5]{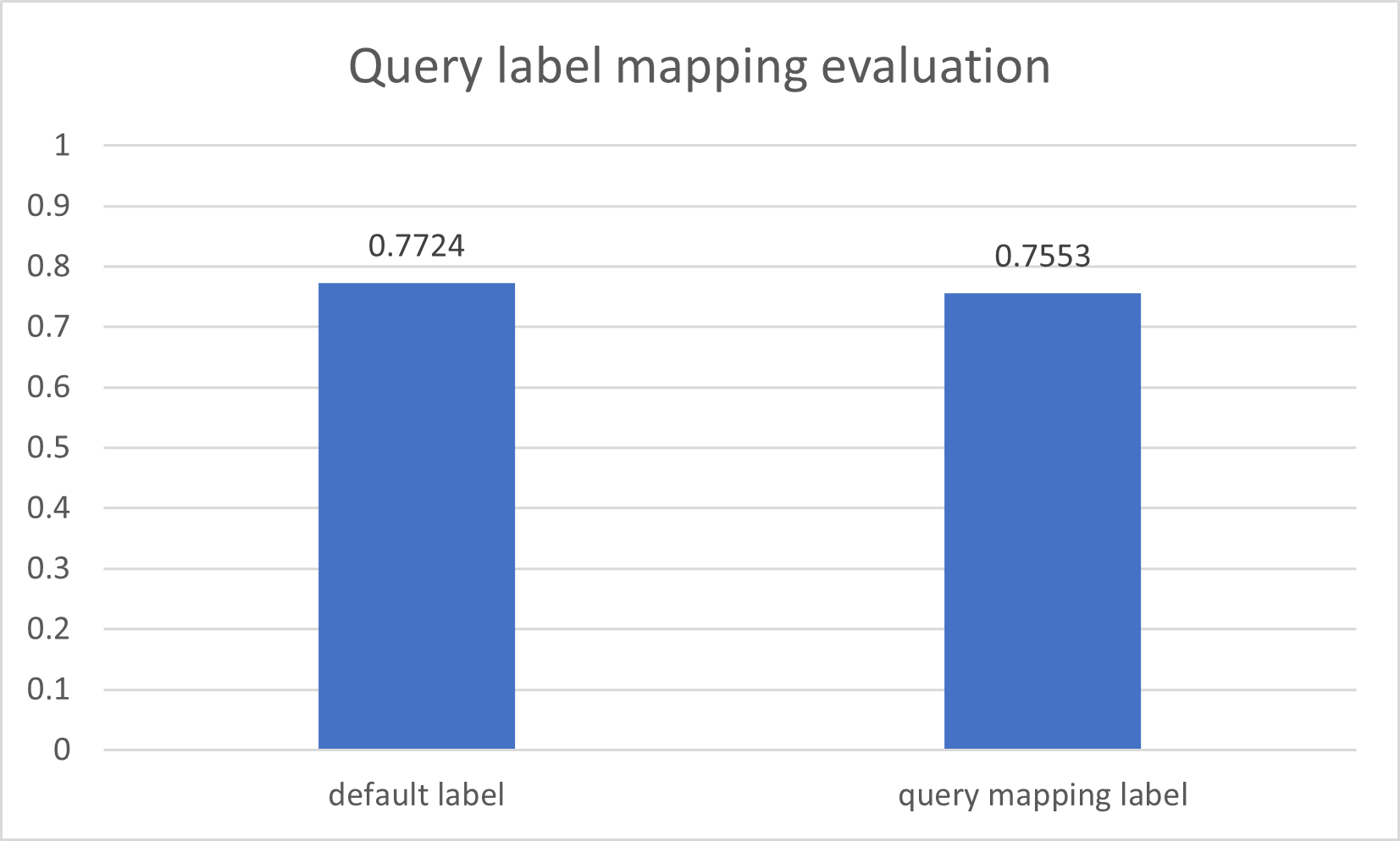}
\caption{\textbf{SBERT query label mapping evaluation.} The figure shows the test accuracy of the original labels and mapping wiki-cate labels of AG News.}
\label{fig-label-mapping-evaluation}
\end{figure}

In the experiment, we compare each text (query) in the AG News test set with all wiki categories using cosine similarity. We obtain the category closest to this query as input (for example, query: \textit{Fears for T N pension after talks Unions representing workers at Turner   Newall say they are 'disappointed' after talks with stricken parent firm Federal Mogul.} corresponds to category: \textit{Transport and General Workers' Union}). We then perform a second round of cosine similarity between this category and the labels that should be output for the test. The experimental results can be referred to the Figure \ref{fig-label-mapping-evaluation}.

\subsection{Self-training}
The self-training experiments can be divided into three parts to assess the impact of their configurations on accuracy:

\begin{itemize}
    \item \(i\): The number of iterations
    \item \(m\): The choice of the model for fine-tuning
    \item \(t\): The setting of the similarity score threshold
\end{itemize}

\begin{figure}[h]
\centering
\includegraphics[scale=0.35]{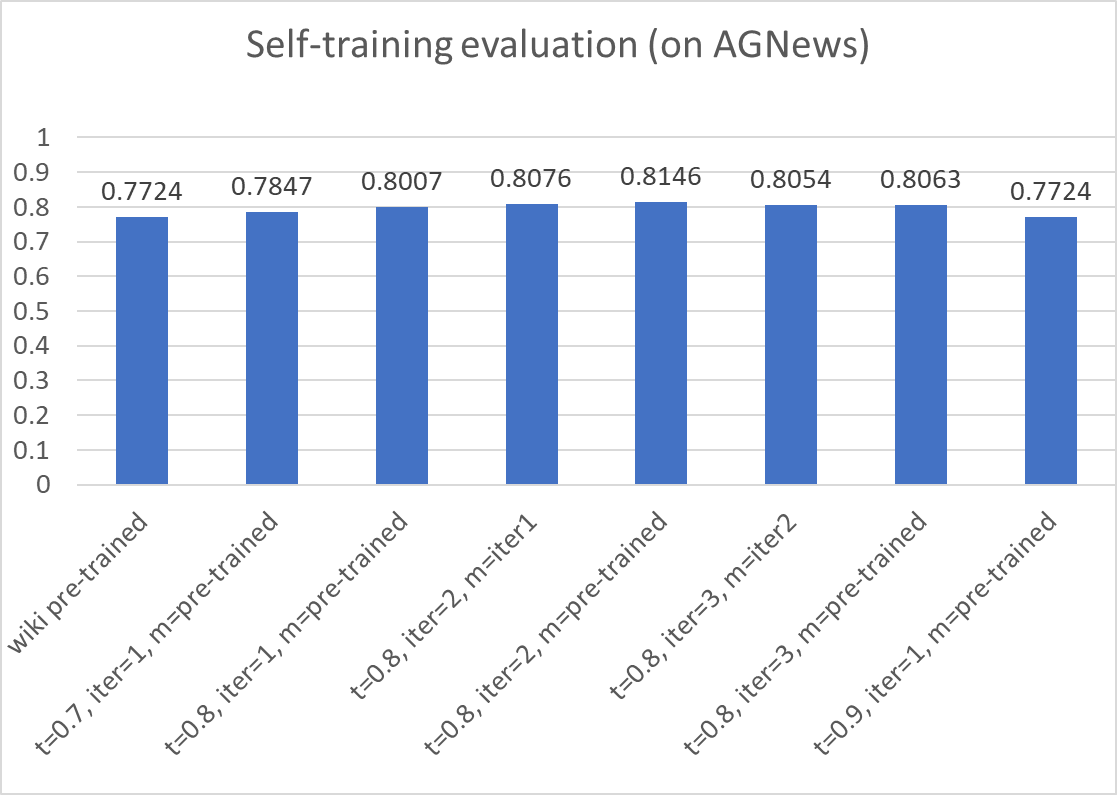}
\caption{\textbf{Self-training evaluation.}}
\label{fig-self-training}
\end{figure}

This study conducted relevant experiments on the AG News test set. Detailed results can be referred to Figure \ref{fig-self-training}.

\subsection{Self-training performance comparison}

Continuing from the previous experiment, Self-training relies on iterative training to increase accuracy. The iterative process consists of two main parts: inference on the target training and testing sets, and fine-tuning based on the inference results. We conducted the experiments using WC-SBERT on different datasets and reproduced the current best model in \citet{selftrain} approach. The experiment involved 2 rounds of fine-tuning and 3 rounds of inference to compare the time cost of inference and fine-tuning between the two methods, thereby evaluating their performance.

To ensure the experiment's fairness, all tests were conducted using the same hardware environment with the following specifications: CPU: Intel(R) Xeon(R) Gold 6154 (8 cores), GPU: NVIDIA Tesla V100 * 2, RAM: 128 GB, OS: Ubuntu 20.04 LTS. The experimental results are presented in Table \ref{speed-comparison}.

\subsection{GPT experiment}

We have also used the currently popular GPT-3.5 model and the GPT-4+fine-tune model provided by the API to test the effects of zero-shot text classification. Compared to the GPT-4 model, the GPT-3.5 model is more affordable, allowing us to test on several datasets. For GPT-3.5, the prompt template we used is as follows:

\begin{listing}[h]%
\caption{Prompt sample with GPT 3.5}
\label{gpt-prompt-listing}%
\begin{lstlisting}
There are 10 classes below:
    Social & Culture
    Science & Mathematics
    Health
    ...
This text is belong to which class: 
Which celebrity smokes the most weed? snoop dog of course
\end{lstlisting}
\end{listing}

Using the aforementioned prompt for inference on the GPT-3.5 model, the results are presented in the experiment result section. 

Moreover, we also tried using the GPT-4 + fine-tune provided by OpenAI API to address the zero-shot text classification problem. However, due to its high cost, we only used 6,700 Wikipedia entries (split into 5,700 training data and 1,000 validation data) to fine-tune the GPT-4 model. We then randomly selected 300 entries from the Yahoo topic as a test set. If we fine-tuned the GPT-4 model with all 5,700 training data entries at once, the resulting test accuracy was 0.55, far from state of the art. However, OpenAI API also offers the feature to fine-tune an already fine-tuned model, meaning you can fine-tune a model a second time with a smaller dataset. By dividing the 5,700 entries into 5,000 for the first round of fine-tuning and 700 for the second round, we preliminarily achieved a test accuracy of 0.62. Of course, due to the high cost of using the Curie model, we could only preliminarily test on a few hundred entries. This GPT fine-tuning approach is also challenging to apply to large-scale datasets for zero-shot text classification.

\subsection{Experiment results}
From the above experiments, several inferences can be drawn:
\begin{itemize}
    \item SBERT-based similarity models, when combined with labels as the pre-training and fine-tuning objective, still show sensitivity to the use of prompts. Using prompts leads to better accuracy.
    \item From the label mapping experiment, we have shattered conventional intuition. Directly comparing the query with the target dataset's labels using SBERT's similarity yields better results than mapping the query first.
    \item Self-training effectively improves accuracy within 1 to 2 iterations.
    \item Compared to the current best model in Gera et al. (2022), WC-SBERT demonstrates significant improvements in both inference and fine-tuning performance. During the inference phase, WC-SBERT achieves a remarkable reduction in time across different datasets, ranging from 10 to 40 times faster. In the fine-tuning phase, Self-Gera only utilizes 800 samples from the target set, whereas WC-SBERT employs 6.45 million samples from WIKI for fine-tuning. Despite the difference in dataset sizes, WC-SBERT's overall fine-tuning time remains lower than that of Self-Gera, resulting in a performance boost of 5,000 to 9,700 times.
    \item GPT-3.5 model does not achieve state-of-the-art performance on several datasets, shows poor results, and has higher costs compared to other models. Currently, it is not suitable for use in zero-shot text classification problems.
\end{itemize}

Based on the above inferences, we adopt this framework as our foundational model architecture and conduct experiments on different target datasets as shown in the Table \ref{table-exp-result}.

\begin{table*}[h]
\scriptsize
\centering
\begin{tabular}{l|ccc|ccc|ccc}
\hline
    \textbf{Dataset} & \multicolumn{3}{c}{\textbf{AG News}}    & \multicolumn{3}{c}{\textbf{DBpedia}} & \multicolumn{3}{c}{\textbf{Yahoo}} \\
\hline
    Model & \multicolumn{1}{c}{Self-Gera} & \multicolumn{1}{c}{WC-SBERT} & \multicolumn{1}{c}{time ratio} & \multicolumn{1}{c}{Self-Gera} & \multicolumn{1}{c}{WC-SBERT} & \multicolumn{1}{c}{time ratio} & \multicolumn{1}{c}{Self-Gera} & \multicolumn{1}{c}{WC-SBERT} & \multicolumn{1}{c}{time ratio}  \\
\hline
Inference (\(i\)=0) & 204 & 20 & 10.2 & 7939 & 304 & 26.12 & 13201                & 300 & 44.00 \\
Inference (\(i\)=1) & 207 & 20 & 10.35 & 8077 & 299 & 27.01 & 11525               & 300 & 38.42 \\
Inference (\(i\)=2) & 206 & 20 & 10.3 & 8197 & 300 & 27.32 & 11493                & 300 & 38.31 \\
Total time (sec.) & 617 & 59 & 10.46 & 24213 & 904 & 26.78 &              36219 & 901 & 40.20 \\
\#sample      & 7600 & 7600 & - & 70000 & 70000 & - & 58966                 & 60000 & - \\
Avg. time (sec.)  & 0.0271 & 0.0026 & \textbf{10.42} & 0.1153                   & 0.0043 & \textbf{26.79} & 0.2047                            & 0.0050 & \textbf{40.92} \\
\hline
Fine-tune (\(i\)=1) & 110 & 91 & 1.21 & 395 & 110 & 3.60 & 221                    & 133 & 1.66 \\
Fine-tune (\(i\)=2) & 106 & 87 & 1.22 & 391 & 124 & 3.15 & 218                    & 134 & 1.63 \\
Total time (sec.)  & 216 & 179 & 1.21 & 785 & 234 & 3.35 & 438                   & 267 & 1.64 \\
\#sample       & 800 & 6458670 & - & 2800 & 6458670 & - 
               & 2000 & 6458670 & - \\
Avg. time (sec./100 samples)   & 13.5039 & 0.0014 & \textbf{9766.34} 
               & 14.0262 & 0.0018 & \textbf{7738.85} 
               & 10.9610 & 0.0021 & \textbf{5219.52} \\
\hline
\end{tabular}
\caption{\textbf{Self-training speed comparison between WC-SBERT and Self-Gera.} We compare the training and inference time between our WC-SBERT and self-training model in \citet{selftrain}. The backbone of the former model is \textit{all-mpnet-base-v2}, while the latter model uses \textit{Narsil/deberta-large-mnli-zero-cls} as its backbone.  \(i\) denotes iteration of self-training, "Inference Avg. time" represents the average time spent per sample data, "Fine-tune Avg. time" represents the average time spent per 100 sample data, and "time ratio" is the time of Self-Gera divided by the time of WC-SBERT. The numerical values presented in the table are rounded to the nearest integer to show the measurement errors.}
\label{speed-comparison}
\end{table*}

\begin{table*}[h]
\scriptsize
\centering
\begin{tabular}{l|l|l|l|l|l|l|l|l|l|l|l|l|l|l|l}
\hline
\textbf{Label index}&0&1&2&3&4&5&6&7&8&9&10&11&12&13&\(A.\) \\
\hline
0 (Company)&3115&80&52&28&23&517&118&60&21&12&113&579&186&96&62.3 \\
1 (EducationInstitution)&14&4768&55&11&7&2&27&16&64&6&7&6&14&3&95.36 \\
2 (Artist)&38&14&1348&11&18&0&38&18&14&25&12&2254&446&764&26.96 \\
3 (Athlete)&7&0&4&4950&1&0&4&1&2&3&0&23&4&1&99.00 \\
4 (OfficeHolder)&260&117&44&54&3757&21&245&56&239&51&14&8&32&102&75.14 \\
5 (MeanOfTransportation)&1088&0&1&17&0&3381&16&480&8&6&1&1&1&0&67.62 \\
6 (Building)&79&239&89&5&3&33&2347&791&1376&11&10&2&9&6&46.94 \\
7 (NaturalPlace)&0&0&0&0&0&2&1&4976&21&0&0&0&0&0&99.52 \\
8 (Village)&0&2&0&0&0&0&0&363&4635&0&0&0&0&0&92.70 \\
9 (Animal)&0&0&0&17&0&0&0&67&0&1972&2944&0&0&0&39.44 \\
10 (Plant)&8&0&0&0&0&1&0&6&0&3&4979&0&3&0&99.58 \\
11 (Album)&0&0&1&1&0&0&0&0&0&0&0&4996&1&1&99.92 \\
12 (Film)&2&0&4&1&0&0&1&3&3&5&0&442&4538&1&90.76 \\
13 (WrittenWork)&264&53&273&5&61&15&91&34&112&208&118&275&1293&2198&43.96 \\
\hline
\end{tabular}
\caption{ \textbf{Error analysis of DBpedia (without description prompts).} The predicted labels by our WC-SBERT are presented as columns, while the actual labels are presented in rows. The rightmost column displays the test accuracy for each label. Results with desciption prompts is shows in Table \ref{table-dbpedia-confusion-matrix-2} in the appendix.}
\label{table-dbpedia-confusion-matrix-1}
\end{table*}

\begin{table}[h]
\small
\centering
\begin{tabular}{l|l|l|l}
\hline
\textbf{Dataset} & \textbf{AG News} & \textbf{Yahoo} & \textbf{DBpedia} \\
\hline
WC-SBERT & \textbf{0.815} & \textbf{0.627} & 0.7481 \\
& (\(i\)=2, \(t\)=0.8) & (\(i\)=1, \(t\)=0.8) & (\(i\)=1, \(t\)=0.7) \\
Wiki-Ding & 0.796 & 0.573 & 0.902 \\
Self-Gera & 0.814 & 0.62 & \textbf{0.945} \\
GPT 3.5 & 0.71 & 0.597 & N/A \\
\hline
\end{tabular}
\caption{\textbf{Evaluation results of different target datasets with prompts.} The numbers denote test accuracy of our model and three baselines: \citet{tewiki}, \citet{selftrain}, and GPT3.5. Due to the cost of GPT-3.5, we do not perform GPT-3.5 inference on the DBPedia test set, which consists of a total of 70,000 samples.}
\label{table-exp-result}
\end{table}

\section{Discussion}

Based on the experimental results, it can be observed that WC-SBERT achieves state-of-the-art performance in zero-shot topic intent classification tasks such as AG News and Yahoo! topic, while significantly reducing the training time required for self-training. However, when facing the DBpedia task, it did not perform well, achieving only 0.74 accuracy. Therefore, we conducted further error analysis specifically for DBpedia in an attempt to understand the reasons affecting accuracy.

Table \ref{table-dbpedia-confusion-matrix-1} shows that certain labels of DBpedia are incorrectly classified. For example, although there should be 5,000 instances classified as \textit{Animal} in the actual data, the prediction results include 2,944 instances misclassified as \textit{Plant}. We believe this situation may be due to the presence of ambiguous definitions in some labels. Based on the definitions and content of the dataset, \textit{Artist} can encompass professionals in various fields, including painters, musicians, and producers. \textit{Album} refers to music-related albums, and \textit{Film} may cover music and other audiovisual content as well. It is understandable that confusion and ambiguity may arise from the label definitions.

In earlier experiments, we realized the importance of prompts on SBERT's performance in classification tasks. Therefore, we defined specific prompt formats for each label in DBpedia: "This {description} described in this content is {label}." We added explicit keywords and self-defined descriptions for each label, aiming to enhance discriminability by distinguishing the characteristics of each label. For example, the description of \textit{Athelete} is "This person who plays sport described in this content is an athlete." Other prompts with descriptions of DBpedia can be referred to Table \ref{table-descri} in the appendix. Using the updated prompts, we performed inference with the same model again, and the accuracy increased from 0.7423 to 0.8746. Although there is still a slight difference in accuracy compared to results achieved by \citeauthor{selftrain}, they requires access to the training and test datasets for downstream fine-tuning tasks. In contrast, our study only use data from Wikipedia, resulting in an improvement of nearly 20\% in accuracy compared to the original 0.74. Based on the results of the error analysis, in addition to SBERT's sensitivity to prompts, the predictive ability for classification tasks can also be enhanced by adding keywords and descriptions to the labels.

We also employ description prompts on both AG News and Yahoo! Topics using the same approach. We incorporated specific descriptions for a few categories. Based on previous experiments, we understand that SBERT utilizes contextual understanding to compare the similarity or distance between two sentences. In an attempt to increase the distinction between two ambiguous categories, we injected reverse implications by using the term "not." For example, in the case of AG News, we modified the original prompt to "This topic is talk about World, not Business." All prompts are presented in Table \ref{table-prompt-agnews} and \ref{table-prompt-yahoo} in the appendix. The accuracy improved as shown in Table \ref{table-prompt-result}; however, further manual adjustments are needed to refine the prompt definition and descriptions for better effectiveness. One of our future works is to explore how to automatically generate appropriate prompts for different datasets.

\begin{table}[h]
\centering
\begin{tabular}{l|l|l|l}
\hline
\textbf{Dataset} & \textbf{AG News} & \textbf{Yahoo} & \textbf{DBpedia} \\
\hline
Default prompt & 0.815 & 0.627 & 0.7481 \\
Description prompt & \textbf{0.837} & \textbf{0.634} & \textbf{0.8746} \\
\hline
\end{tabular}
\caption{\textbf{Test set accuracy of the default and description prompts.}}
\label{table-prompt-result}
\end{table}

\section{Conclusion}
This paper introduces a novel model for zero-shot text classification that is adaptable to open-domain datasets. Traditional methods typically involve extensive training on large amounts of text. In our approach, we propose WC-SBERT, a combination of SBERT with the Wikipedia dataset, and utilize self-training to train the Wikipedia categories. This approach significantly improves training efficiency and reduces computing time. Our experimental results demonstrate that WC-SBERT substantially decreases the training and inference time per data point and achieves state-of-the-art performance on popular text classification datasets, such as AG News and Yahoo.

\bibliography{aaai24}

\clearpage
\section{Appendix}

\begin{table}[h]
\centering
\begin{tabular}{l|l}
\hline
\textbf{Label} & \textbf{Description prompt} \\
\hline
World&This topic is talk about World not Business \\
Sports&This topic is talk about Sports \\
Business&This topic is talk about Science not World \\
Science&This topic is talk about Science \\
Technology&This topic is talk about Technology \\
\hline
\end{tabular}
\caption{\textbf{Description prompts of labels on AG News.}}
\label{table-prompt-agnews}
\end{table}

\begin{table}[h]
\centering
\begin{tabular}{l|l}
\hline
\textbf{Label} & \textbf{Description prompt} \\
\hline
Society&This topic is talk about Society not Family or Relationships \\
Education&This topic is talk about Education not Science or Mathematics \\
\hline
\end{tabular}
\caption{\textbf{Description prompts of labels on Yahoo! Topic.} We only provided descriptions for a few categories in our analysis.}
\label{table-prompt-yahoo}
\end{table}

\begin{table*}[h]
\small
\centering
\begin{tabular}{l|l}
\hline
\textbf{Label} & \textbf{Description prompt} \\
\hline
Company&This topic is describing this company \\
EducationInstitution&This school, university described in this content is an education institution \\
Artist&This musician, painter, singer, writer, author described in this content is an artist \\
Athlete&This person who plays sport described in this content is an athlete \\
OfficeHolder&This person who holds a position or office in a government described in this content is an officeholder \\
MeanOfTransportation&This vehicles, ridden, trains and other conveyances described in this content is transportation \\
Building&This man-made structure described in this content is a building \\
NaturalPlace&
\begin{tabular}[c]{@{}l@{}}This natural landforms, bodies of water, vegetation, rocks, forests, rivers, lakes, mountains, oceans, grasslands \\ described in this content is a natural place\end{tabular} \\
Village&This town, small settlement or community described in this content is a village \\
Animal&This organism described in this content is an animal \\
Plant&This organism described in this content is a plant \\
Album&This music or recorded tracks described in this content is an album \\
Film&This movie described in this content is a film \\
WrittenWork&This books, essays, poems or literatures described in this content is a written work \\
\hline
\end{tabular}
\caption{\textbf{Description prompts of labels on DBpedia.}}
\label{table-descri}
\end{table*}

\begin{table*}[h]
\scriptsize
\centering
\begin{tabular}{l|l|l|l|l|l|l|l|l|l|l|l|l|l|l|l}
\hline
\textbf{Label index}&0&1&2&3&4&5&6&7&8&9&10&11&12&13&\(A.\) \\
\hline
0 (Company)&3096&71&87&6&19&607&224&18&6&15&60&464&213&114&61.92 \\
1 (EducationInstitution)&28&4804&58&1&6&1&51&3&31&2&3&0&9&3&96.08 \\
2 (Artist)&61&22&2862&5&37&1&113&3&0&10&4&710&235&937&57.24 \\
3 (Athlete)&8&1&26&4941&4&4&7&0&0&1&0&1&6&1&98.82 \\
4 (OfficeHolder)&100&92&80&17&4441&10&157&2&16&8&1&2&5&69&88.82 \\
5 (MeanOfTransportation)&75&0&0&0&0&4853&58&8&0&3&0&1&1&1&97.06 \\
6 (Building)&91&208&35&0&2&55&4262&110&207&8&1&1&7&13&85.24 \\
7 (NaturalPlace)&0&0&1&0&0&0&45&4853&101&0&0&0&0&0&97.06 \\
8 (Village)&0&5&1&0&0&0&12&92&4890&0&0&0&0&0&97.80 \\
9 (Animal)&0&0&2&6&0&0&0&186&0&4423&382&0&0&1&88.46 \\
10 (Plant)&8&0&0&0&0&2&1&46&0&11&4931&0&1&0&98.62 \\
11 (Album)&0&0&5&1&0&0&0&0&0&0&0&4992&1&1&99.84 \\
12 (Film)&4&0&26&0&0&2&3&0&0&2&0&326&4629&8&92.58 \\
13 (WrittenWork)&584&50&87&0&9&14&32&12&16&144&29&132&645&3246&64.92 \\
\hline
\end{tabular}
\caption{\textbf{Error analysis of DBpedia (with description prompt).} The predicted labels are presented as columns, while the actual labels are presented in rows. The rightmost column displays the test accuracy for each label.}
\label{table-dbpedia-confusion-matrix-2}
\end{table*}

\end{document}